\documentclass[conference]{IEEEtran}
\IEEEoverridecommandlockouts
\usepackage{cite}
\usepackage{amsmath,amssymb,amsfonts}
\usepackage{algorithmic}
\usepackage{graphicx}
\usepackage{textcomp}
\usepackage{xcolor}

\begin{document}

\title{A Brief Survey of Machine Learning Methods for Emotion Prediction using Physiological Data} 

\author{
	\IEEEauthorblockN{ Maryam Khalid}
	\IEEEauthorblockA{		\textit{Rice University}\\
		Houston, USA \\
		mk79@rice.edu}
	\and
	\IEEEauthorblockN{Emily Willis}
	\IEEEauthorblockA{\textit{Rice University}\\
		Houston, USA \\
	emw15@rice.edu}
}

\maketitle

\begin{abstract}
	
Emotion prediction is a key emerging research area that focuses on identifying and forecasting the emotional state of a human from multiple modalities. Among other data sources, physiological data can serve as an indicator for emotions with an added advantage  that it cannot be masked/tampered by the individual and can be easily collected. This paper surveys multiple machine learning methods that deploy smartphone and physiological data to predict emotions in real-time, using self-reported ecological momentary assessments (EMA) scores as ground-truth. Comparing regression, long short-term memory (LSTM) networks, convolutional neural networks (CNN), reinforcement online learning (ROL), and deep belief networks (DBN), we showcase the variability of machine learning methods employed to achieve accurate emotion prediction. 

We compare the state-of-the-art methods and highlight that experimental performance is still not very good. The performance can be improved in future works by considering the following issues: improving scalability and  generalizability, synchronizing multimodal data, optimizing EMA sampling, integrating adaptability with sequence prediction, collecting unbiased data, and leveraging sophisticated feature engineering techniques.

\end{abstract}

\begin{IEEEkeywords}
Emotion regulation, physiological, Machine learning models, EMA, Adaptive models, spatiotemporal data, multimodal 
\end{IEEEkeywords}

\section{Introduction}
Emotion recognition (ER) is a sub-field of affective computing that was proposed by Rosalind Picard in 1997 \cite{picard}. Affective computing, derived from the word 'affect' or 'emotion', is  computing that relates to, arises
from, or influences emotions. The field spans quantification, modeling, estimation and prediction of human emotional state, and  integrating it into a computer system  to enable the machine to make \textit{affect-aware} decisions. In this paper, we will conduct a literature review of emotion prediction only. This paper will not only provide an in-depth survey of the state-of-the-art methods for emotion prediction but also serve as a primer on this topic.

\subsection{What is emotion?}
Paul Ekman identified seven universal emotions in terms of which all other compound emotions can be defined: happiness, sadness, surprise, fear, anger, disgust, and contempt \cite{emotion}. Please note that  these are all abstract terms, and it is a huge challenge to quantify them. Each individual feels emotions  in a different way and currently there's no central metric to calibrate each individual's emotion to a standard scale. Many models have been proposed in literature to quantify emotion. Two main models/techniques that we encountered in literature on emotion prediction are Geneva Emotion Wheel (GEW)\cite{geneva} and Ecological Momentary Assessments (EMA).

\begin{figure}
	\includegraphics[scale=0.37]{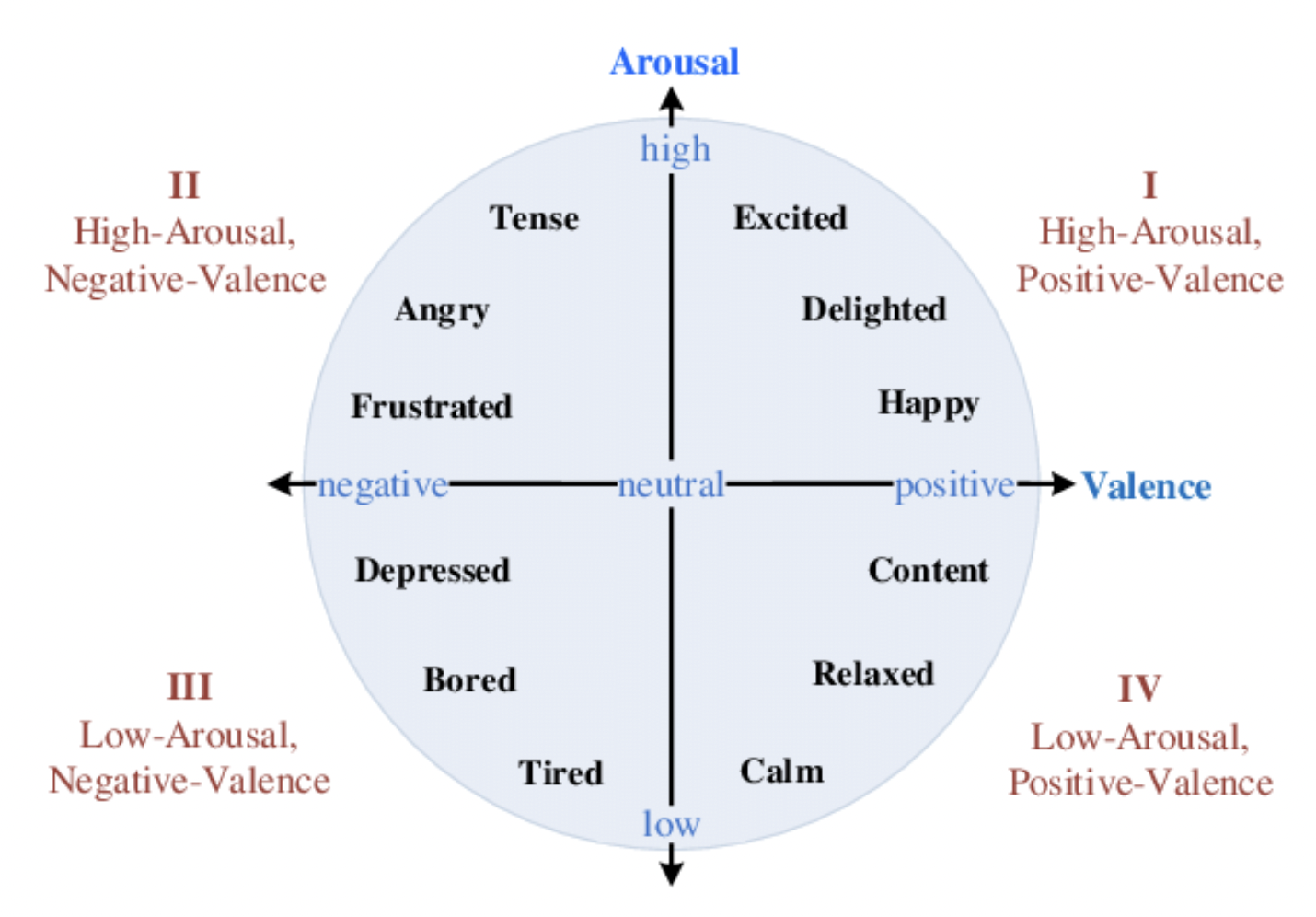}
	\caption{Geneva Emotion Wheel \cite{gfigure}}
	\label{fig:1}
\end{figure}

GEW explains emotions in a 2D space defined by valence on one axis and arousal on the other, as shown in Fig. \ref{fig:1}. Theoretically, arousal and valence both take continuous values. However, since these values are determined by users, in most experiments these scales are discretized uniformly between 1 and 10.

Ecological momentary assessment (EMA) scores are used as a measure of the dimensions of mental health in the moment that the survey is recorded. For the CrossCheck dataset in particular, it serves as ground truth for the prevalence of schizophrenia symptoms. The survey used for that particular  dataset is comprised of 10 one-sentence questions about the current wellbeing of the participant to which they reply a number 0-3 based on the dimension of that aspect of their wellbeing.The number and type of questions are a design choice.  This choice trades off between user's convenience and amount of information.

\subsection{How is emotion measured?}
After modeling the emotion, the next important question is how it can be measured in terms of more quantifiable and easily accessible data sources. This is where learning comes into the picture and the active research transitions from psychology to  
engineering. Before we discuss the learning methods, it is important to discuss sources of data that are utilized to learn about emotion. \\
There are two main categories of data: non-physiological and physiological. The former mainly is comprised of two  subcategories. The first one is facial expression, text, and speech, which are captured through images and video. The second subcategory is movement and gesture, which could be captured by IMU sensors and GPS. We grouped these signals together under of the umbrella of non-physiological signals because the user has a control over them and  can modify/mask them. For example, a person can be angry but still be able  to control their facial expressions in order  to not show it. However, they can not control their physiological signals, which are comprised of the following:
\begin{itemize}
	\item  Electroencephalography (EEG), which measures brain activities
	\item Electrodermal Activity (EDA), which measures electrical conductivity or skin conductance
	\item Electrooculogram (EOG), which  measures eye movements
	\item Electrocardiogram (ECG), which  measures heart beats
\end{itemize}
In this paper, we will focus  on physiological signals only. 

\section{Machine Learning Methods}
After briefly describing emotion and its data sources, now we will discuss the common machine learning methods that are used for emotion prediction. Some methods, like reinforcement online learning and deep belief networks, were not discussed in class, so a small introduction is provided before discussing them in the context of the emotion prediction problem.

\subsection{Regression}
Regression analysis is commonly used among many of the reviewed papers. For instance, Tseng et al. utilizes multi-output support vector regression (SVR) machines to train their models and compare their performance with the performance of other models trained using multi-task learning (MTL) and single-task learning (STL) \cite{R1}. This algorithm works similarly to the optimization of single-output SVR; multi-output SVR learns the mean regressor $w_0$ for all tasks $1, ..., T$ and finds small vectors $v_1, ...,v_T$ that adjust the mean regressor to account for the task relatedness while predicting the output of each task and minimizing the overall prediction error\cite{R1}. Yu and Sano also use regression models to predict the next day’s mood, health, and stress scores and then evaluate the performance of the various algorithms used based on these regression models \cite{R3}. In a previous paper, Yu et al. also predict next day’s wellbeing but instead by developing personalized, regularized MTL regression models\cite{R9}. Additionally, Wang et al. runs bivariate regression models before running a prediction analysis using gradient boosted regression trees (GBRT) \cite{R2}. The purpose of regression in this paper is to analyze the self-reported EMA scores and their association with the passive sensor data as well as to better understand how practical it is to predict these scores. Lastly, Jaques et al. utilize Gaussian process regression as the base model for domain adaptation in an attempt to obtain a very personalized prediction of emotion and wellbeing \cite{R10}. The popularity of this method is a testament to its usefulness when it comes to prediction, understanding the correlation between variables, and assessing the performance of the model.

\subsection{Convolution Neural Network}
In recent work\cite{CNN}, CNN has been proposed for emotion recognition. The input data, like other methods, is composed of ECG signals (32 channels) and peripheral signals (temperature, etc.). There are two models proposed in this work. The first model only takes ECG signals as input, and the architecture is composed of multiple 2D-CNN layers with maxpooling followed by dropout layer,  fully-connected MLPs, and a softmax at the end. The second model takes both ECG and peripheral data as input through multiple independent channels followed by concatenation and MLP. All activations are ReLu, except the last layer which is softmax. The loss function is cross-entropy and the ADAM optimizer is used.\\
The output/labels are  valence and arousal, but they are binned to form 3 classes for each category. They also fuse multiple modalities together but do not report the results because it performs worst that a single modality model.

\subsection{Long Short-term Memory Networks}
Long short-term memory (LSTM) networks are an extension of recurrent neural networks. They can process temporal data that comes in sequences and are able to learn long-term dependencies, so they are useful for prediction when working with time-series data. Yu et al. apply this method for wellbeing prediction \cite{R9}, and then, in a later paper, Yu and Sano combine a LSTM network with a convolutional neural network (CNN) and compare its performance with that of a deep LSTM network \cite{R3}. A LSTM network is a participant-independent, generalized method that works well with temporal data and can be employed to try to achieve a predictive model that adapts well to new participants.

\subsection{Reinforcement Online Learning}
The work presented in \cite{ROL} combines online learning with reinforcement learning. Since these methods were not discussed in class, we  first present a brief summary. As opposed to conventional batch training with the entire training data, online learning trains the model with one sample at a time either  because the data is sequential or is not available completely at a certain time, e.g streaming data \cite{ML1}. Online learning allows the model to adapt as new data comes in, and the model is refined over time compared to one-time batch training with fixed computation complexity. Reinforcement Learning (RL) defines a notion of agent that learns the optimum set of actions, also known as policy, by interacting with its environment. The distinguishing aspect of RL is the notion of reward that drives  the agent towards learning the optimum policy. For each action, a reward is defined, and the agent chooses the next action by maximizing the expected value of future reward.\\
In this work, the classification problem is formulated as minimization of a loss function with  \textit{$l1$} regularization in a stochastic gradient descent setting to find optimum weights of linear classifier. Secondly, for RL problem formulation, each sample or all data points at time step $t$ represent the state at time t. After using the data up until time $t$ to find weights in online ML setting, the model is fitted to the next data point $x_{t+1}$. The estimated label $\hat{y_{t+1}}$ is compared with true label $y_{t+1}$ to determine the reward:

\begin{equation}
	R =
	\begin{cases}
		1 & \text{if   $    ||\hat{y_{t+1}}-y_{t+1}||<\sigma$}\\
		0 & \text{otherwise}
	\end{cases}       
\end{equation}

When the reward is $1$, the model is kept as is. However, when it is $0$, implying it performed poorly, the model is updated. Thus, in that scenario, the online ML training routine is executed using $s_{t+1}$.
The linear classifiers used to evaluate the proposed method include least squares (LS) and support vector regression machine (SVR).

\subsection{Feature Selection: Deep Belief Networks}
The work in  \cite{NN} proposes using Deep Belief Networks (DBN) for feature selection on EEG data to predict arousal, valence, and liking of a user to a one-minute video. The authors argue that hand-picked features by experts or feature selection using PCA is labor-intensive and not scalable, whereas using an automated mechanism to extract features from raw EEG data using DBN is scalable and  adaptive when physiological data or the task changes. However, it is important to note that the authors do not use PCA as a baseline in their performance evaluation, and this claim about PCA scalability is not convincing since ER  datasets are usually  small in size because of limitations involved in collection of survey data from actual humans.

\begin{figure}
	\includegraphics[scale=0.4]{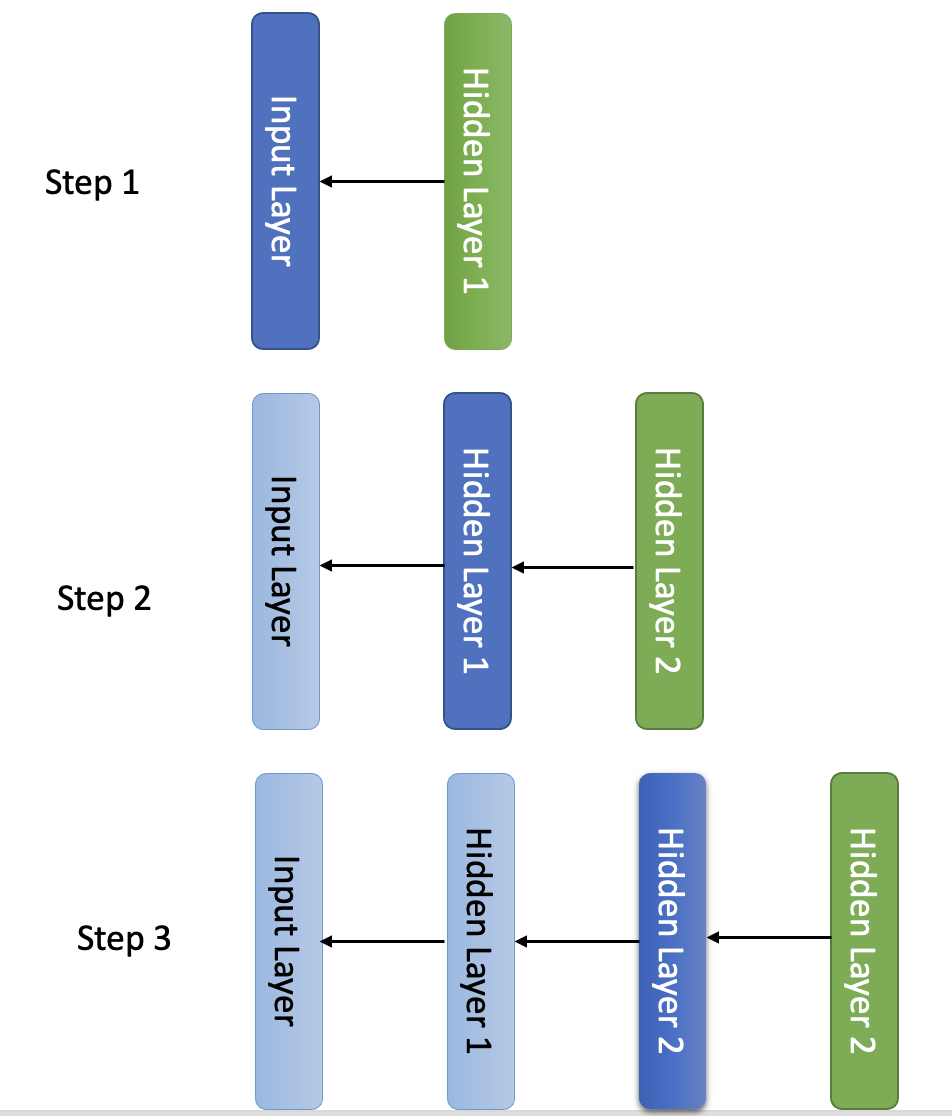}
	\caption{Step-wise  training in DBN}
	\label{fig:dbn}
\end{figure}

Deep Belief Networks are a semi-supervised class of deep neural networks  with multiple hidden layers\cite{DBN1}. The layers are connected, but the hidden units in one layer are not. When fed  unlabeled data, they provide feature selection, and when labels are provided, they can be used for classification. The main difference between neural networks and DBN is that DBN is composed of Restricted Boltzmann Machines (RBM) as the building blocks. An RBM consists of one visible and one hidden layer with no connections between nodes of same layer. An important property of RBMs is that the hidden units are conditionally independent given the visible states. As shown in Fig \ref{fig:dbn}, the network is trained in steps, with one component at each step from bottom-up (or left-right in the figure). At step 1, Input Layer acts as visible layer and Hidden Layer 1 (H1) parameters are learned. In step 2, H1 acts as visible layer and Hidden Layer 2 (H2) parameters are learned. At step 3, H2 acts as visible layer and layer 3 is trained. For supervised classification tasks, a softmax layer is added at the end, and the weights are updated using conventional back-propagation.\\
The work in \cite{NN} utilizes DBN to first extract features from raw EEG, EOG, and EMG data in  a unsupervised fashion. After feature selection, the model is trained using backpropagation to classify data for arousal. Separate binary classification models are trained for valence and liking of video. It must be noted that the signals are not truly raw signals because they are pre-processed using notch and band-pass filters for noise reduction. The performance and experiment details are discussed in section \ref{exp}.

\subsection{Multi-task Learning}
Multi-task learning (MTL) is a method frequently used in a majority of the papers reviewed for this project in conjunction with the other models already mentioned. This technique allows for similarities and differences between tasks to be accounted for while, at the same time, solving multiple tasks. For the sake of space, this method is only mentioned briefly here and its applications  and performance will not be discussed in depth.

\section{Datasets}\label{dt}
Physiological data for emotion regulation (ER) is hard to collect for two main reasons. Firstly, the ground truth is very difficult to determine. There is no sensor that can directly measure happiness or sadness. Therefore, we have to rely on an individual's judgment of their emotion,  which leads us to the second challenge associated with this problem: it is a user-centric study. Since the experiments require actual reporting from a human subject, the data collection process is tedious and expensive. For the same reason, most (ER) datasets are relatively small and involve no more than 30-50 participants at a given time. On the other hand, the physiological  signals are collected through multiple channels  at high frequency and are, therefore,  high-dimensional. Additionally many physiological signals like EEG require a lab-setting. Unlike wearable data (like skin temperature), lab-setting experiments are hard to conduct and can not be scaled to large population and longer time periods.\\
 Because of all these challenges, most researchers utilize the small number of open-source datasets available online. All the papers  we read for this project utilized one of the 3 datasets: DEAP, SNAPSHOT, and  CrossCheck. We present a brief summary of these datasets below for two main reasons. Firstly they are used in almost all of the ER research. Instead of discussing them separately for each  method/paper, we present them here to void repetition in interest of space.  Secondly, they will provide the reader an insight into ER data and its characteristics based on which the reader can decide which ML methods would be most suitable to overcome the challenges in this domain. We discuss each dataset in context of 3 properties:
 \begin{itemize}
 	\item Curse of dimensionality
 	\item Heterogeneity in  sources
 	\item How is ground truth determined?
 \end{itemize}
\begin{figure}[b!]
	\centering
	\includegraphics[scale=0.25]{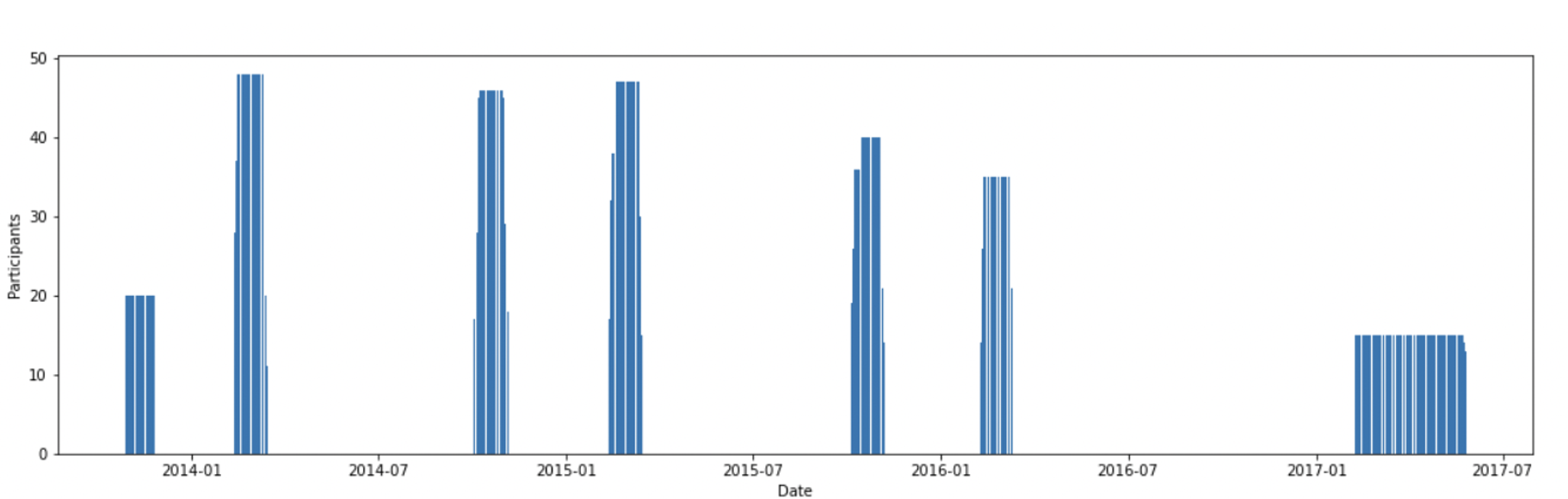}
	\caption{Distribution of participants and length of study periods for SNAPSHOT Data}
	\label{fig:snap}
\end{figure}
\subsection{DEAP}
This multimodal dataset was collected for 32 participants as they watched 40 one-minute long music videos. While the participants were watching the videos, EEG, face video, and physiological signals, including skin resistance, skin conductance, temperature, and EOG, were collected at high frequency. After watching the video, the participants completed an online self-assessment  form rating the videos based on arousal, valence, and dominance, on a scale of 1-9 which is used as ground truth in most papers. After pre-processing and downsampling, the feature matrix X is 40x40x8064, where 8064 is the number of features, 40 is the number of channels (32 for EEG and 9 for other physiological), and  40 is the number of videos/trials. The label for each trial/video is a 4x1 vector with ratings for arousal, valence, dominance, and liking. It can be observed that $p>>n$, and, therefore, this dataset suffers from the curse of dimensionality. Furthermore, data is multimodal and, thus, coming from heterogeneous sources. Each sensor might experience different noise. The labels take continuous values from 1-9, so this dataset can be used for both regression and classification.

\subsection{SNAPSHOT}
SNAPSHOT is also a multimodal dataset\cite{SNAPSHOT} collected by researchers at MIT from 2013-2017  with college students as participants. The  study was conducted over multiple different time periods (avg 2 months each) during each academic term for 4 years.  Different students were recruited in each academic term and the distribution is shown in figure \ref{fig:snap}. Four different types of data were collected: mobile data, physiological data, survey, and weather data. For mobile data, call logs, sms logs, GPS, and screen time were recorded. From wearables, the physiological data is comprised of electrodermal activity (EDA) measured as skin conductance (SC), skin
temperature (ST), and 3-axis acceleration (AC) collected at high frequency. Online surveys were filled by participants each morning and evening and contained information about drugs intake, alcohol, sleep time, naps, exercise, and academic information. For ground truth emotional state,  the following EMAs were filled by each participant : State Anxiety Score, Trait Anxiety Score, Physical Component Score, and Big 5 Personality Test.

Altogether, there were around 240 participants in the study. The physiological data WAS collected at high frequency, so it is downsampled by taking mean and variance for each day (since labels are available for each  morning and evening). After pre-processing, there are around 400 features. If each academic term is used separately, then it becomes a $p>n$ dataset that suffers from curse of dimensionality since the number of participants at any given time are small. The problem is reduced if all data from 2014 to 2017 is integrated together. Data is highly heterogeneous and many sources, such as GPS and wearables, suffer from high noise.
The ground truth is collected on a scale of 1-100, so it can be used for both regression and classification.
\begin{figure}[t!]
	\centering
	\includegraphics[scale=0.4]{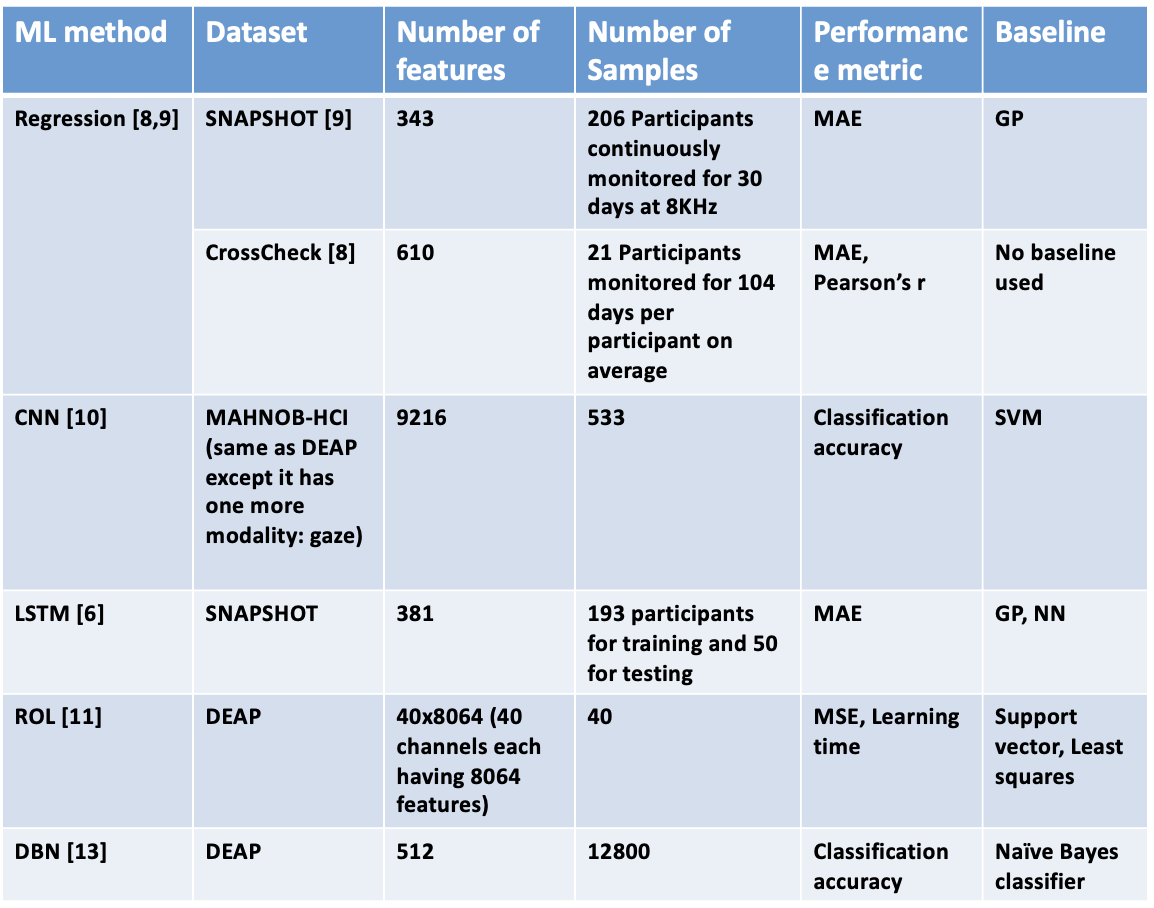}
	\caption{Experiment settings }
\end{figure}

\subsection{CrossCheck}
The CrossCheck dataset was collected from 61 participants by the CrossCheck system over the course of 6,132 days \cite{R1}. This system continuously and passively collected raw sensor data to record both the behavioral and environment rhythms of the participants, including data on app usage, acceleration, calls and SMS, screen on/off activity, location, ambient environment, speech and conversation, and sleep \cite{R1}\cite{R2}. Additionally, every 2-3 days the system prompts users to report their EMA scores, which serve as ground truth. This high-dimensional dataset is heterogeneous, given the heterogeneity of both patients and schizophrenia symptoms, as well as multivariate. Due to the nature of how the data was collected, the data must be preprocessed before use. Data cleaning is an important part of this process due to some participants not being a part of the study for a long enough period of time. Additionally, there is missing data due to broken sensors or sensors not being charged, resulting in days with little data collected. Data cleaning reduces noise present as a result of any missing data. Tseng et al. also extracted rhythm features from the passive sensor data while preprocessing the data\cite{R1}.

\begin{figure*}[h]
	\includegraphics[width=\textwidth,height=12cm]{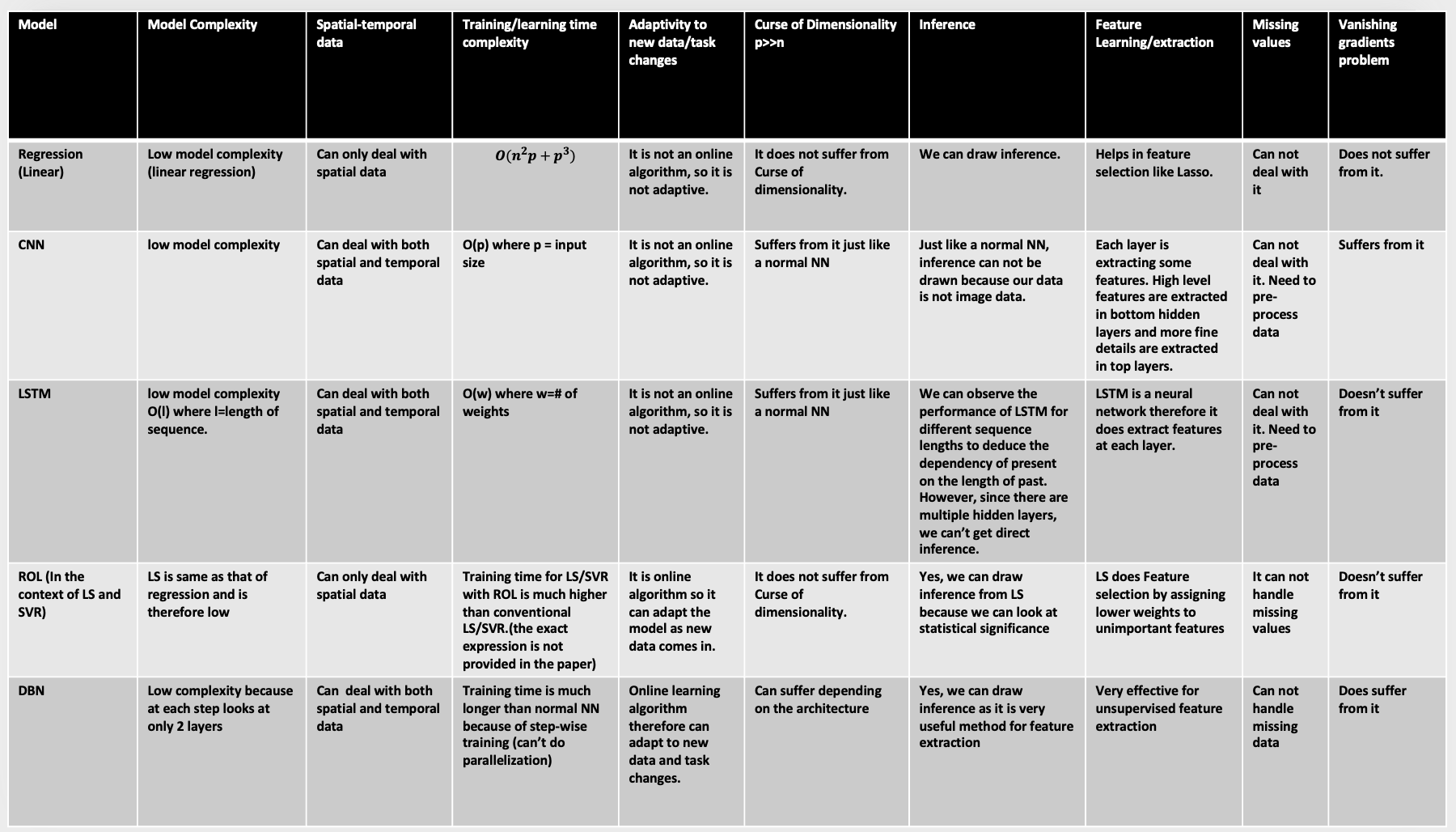}
	\caption{Comparison table}
	\label{fig:tt}
\end{figure*}

\section{Experimental Evaluation}\label{exp}

The purpose of this section is to provide a birds-eye view of experimental evaluation of all methods. Since it was not feasible for us to do the non-trivial data preprocessing and  implement all these methods by ourselves, we report some interesting results for the original papers. Please note that while some papers use the same dataset, their preprocessing and experiment settings are very different, and therefore,  we cannot draw a comparison between reported performances. \\
Before moving to the results, there are some comments about experiment settings. First of all, Mean Absolute Error (MAE) is not the best metric because it only shows absolute error about which we can not conclude whether it is high or low. Compared to MAE,  normalized scores are more informative (indicating whether the model predicted certain emotional states or not). Secondly, along with prediction accuracy,  false positive rate and false negative rate can also provide more insights. Please also note that in most papers,  hyperparameter tuning was either not conducted or not reported in the paper.\\
The prediction accuracy for DBN and CNN is around $60\%$ which is slightly higher than the corresponding baselines but is slight not much better than random ($50\%$). For the remaining methods, the metric is MAE and its hard to infer whether its a large error or small. However, all of them are performing better than their baselines.

\section{Comparison between different methods}

The comparison between different methods is provided in Fig. \ref{fig:tt}. The crux of this analysis is that the choice of model should be determined base don the characteristics of the available dataset. If the data is spatiotemporal, we should deploy LSTM, as it learns from sequences and can handle the vanishing gradient problem. Secondly, the specific goal of the experiment also plays a crucial role in the selection of model. If we are doing offline one-time experiment/evaluation to predict emotion, then regression, CNN, and LSTM (temporal data) are all reasonable choices. However, if, in addition to predicting emotion, an intervention (recommendation to regulate emotion)  is also required, then an online method should be deployed. This experiment would encounter  very dynamic data due to change in environment and circumstances. Therefore, online methods such as ROL and DBN, are more effective. Finally, we should consider the fact that the datasets in this domain have a small number of samples (especially labels) because collecting EMA data is expensive and time-consuming. Therefore, low complexity models should be preferred.

\section{Challenges and Future Work}
\subsection{Feature Engineering} In most papers, feature selection is not performed. Furthermore, most physiological data is collected at high resolution (order of kHz). However, since labels are not available at the same resolution, the data is downsampled by  a huge scale using simple methods like taking mean or variance. This results in a huge loss of information. In the future, more advanced feature engineering methods and encoding schemes can be deployed to retain this information and improve the performance.

\subsection{Dataset Bias} It can be observed from section \ref{dt} that most studies are conducted with only college students , which introduces a huge bias. In future works, more datasets from more diverse population sections should be collected to reduce this bias.

\subsection{EMA Sampling}
The frequency at which surveys are conducted in order to collect ground truth and some feature data (like SNAPSHOT data) is critical. A higher frequency would frustrate the user, and they might give up filling out the surveys. A low frequency is detrimental for the model's learning process. Furthermore, sometimes the user can be in a situation, like driving, when he can't answer the survey. These are all potential issues that can be addressed in future works with solution in the direction  of non-uniform and more intelligent sampling.

\subsection{Combining adaptability with sequence prediction}
The works that we saw were either utilizing spatial data only in an adaptive setting (ROL) or utilizing spatiotemporal data in sequence prediction model like LSTM. In future, the adaptive setting can be extended to integrate spatiotemporal data.

\subsection{Scalability and Generalizability}
As observed in section \ref{dt}, most datasets are collected in either a lab-scale setting or in a specific (college) setting. It is important to ensure that the proposed methods are generalizable to real-life settings. Furthermore, the current datasets have an extremely limited number of users. Scalability of methods to larger populations is also an issue that should be kept in mind while designing future solutions.

\subsection{Multimodal data challenges}
Since emotion is depicted by a huge number of modalities, some modalities (like the ones obtained from wearables)  are collected at a much higher resolution than other. Alignment and synchronization of multiple modalities and their fusion is a key challenge.

\bibliographystyle{IEEEtran}
\bibliography{A_brief_survey}

\begin{thebibliography}{10}
\providecommand{\url}[1]{#1}
\csname url@samestyle\endcsname
\providecommand{\newblock}{\relax}
\providecommand{\bibinfo}[2]{#2}
\providecommand{\BIBentrySTDinterwordspacing}{\spaceskip=0pt\relax}
\providecommand{\BIBentryALTinterwordstretchfactor}{4}
\providecommand{\BIBentryALTinterwordspacing}{\spaceskip=\fontdimen2\font plus
\BIBentryALTinterwordstretchfactor\fontdimen3\font minus
  \fontdimen4\font\relax}
\providecommand{\BIBforeignlanguage}[2]{{%
\expandafter\ifx\csname l@#1\endcsname\relax
\typeout{** WARNING: IEEEtran.bst: No hyphenation pattern has been}%
\typeout{** loaded for the language `#1'. Using the pattern for}%
\typeout{** the default language instead.}%
\else
\language=\csname l@#1\endcsname
\fi
#2}}
\providecommand{\BIBdecl}{\relax}
\BIBdecl

\bibitem{picard}
R.~W. Picard, ``Affective computing, 1997,'' \emph{Google Scholar Google
  Scholar Digital Library Digital Library}, 1997.

\bibitem{emotion}
\BIBentryALTinterwordspacing
P.~Ekman, ``Are there basic emotions?'' \emph{Psychological review}, vol.~99,
  no.~3, p. 550—553, July 1992. [Online]. Available:
  \url{https://doi.org/10.1037/0033-295x.99.3.550}
\BIBentrySTDinterwordspacing

\bibitem{geneva}
\BIBentryALTinterwordspacing
K.~R. Scherer, ``What are emotions? and how can they be measured?''
  \emph{Social Science Information}, vol.~44, no.~4, pp. 695--729, 2005.
  [Online]. Available: \url{https://doi.org/10.1177/0539018405058216}
\BIBentrySTDinterwordspacing

\bibitem{gfigure}
L.-C. Yu, L.-H. Lee, S.~Hao, J.~Wang, Y.~He, J.~Hu, K.~Lai, and X.~Zhang,
  ``Building chinese affective resources in valence-arousal dimensions,'' 06
  2016.

\bibitem{R1}
V.~W.-S. Tseng, A.~Sano, D.~Ben-Zeev, R.~Brian, A.~T. Campbell, M.~Hauser,
  J.~M. Kane, E.~A. Scherer, R.~Wang, W.~Wang \emph{et~al.}, ``Using behavioral
  rhythms and multi-task learning to predict fine-grained symptoms of
  schizophrenia,'' \emph{Scientific reports}, vol.~10, no.~1, pp. 1--17, 2020.

\bibitem{R3}
H.~Yu and A.~Sano, ``Passive sensor data based future mood, health, and stress
  prediction: User adaptation using deep learning,'' in \emph{2020 42nd Annual
  International Conference of the IEEE Engineering in Medicine \& Biology
  Society (EMBC)}.\hskip 1em plus 0.5em minus 0.4em\relax IEEE, 2020, pp.
  5884--5887.

\bibitem{R9}
H.~Yu, E.~B. Klerman, R.~W. Picard, and A.~Sano, ``Personalized wellbeing
  prediction using behavioral, physiological and weather data,'' in \emph{2019
  IEEE EMBS International Conference on Biomedical \& Health Informatics
  (BHI)}.\hskip 1em plus 0.5em minus 0.4em\relax IEEE, 2019, pp. 1--4.

\bibitem{R2}
\BIBentryALTinterwordspacing
R.~Wang, M.~S.~H. Aung, S.~Abdullah, R.~Brian, A.~T. Campbell, T.~Choudhury,
  M.~Hauser, J.~Kane, M.~Merrill, E.~A. Scherer, V.~W.~S. Tseng, and
  D.~Ben-Zeev, ``Crosscheck: Toward passive sensing and detection of mental
  health changes in people with schizophrenia,'' ser. UbiComp '16.\hskip 1em
  plus 0.5em minus 0.4em\relax New York, NY, USA: Association for Computing
  Machinery, 2016, p. 886–897. [Online]. Available:
  \url{https://doi.org/10.1145/2971648.2971740}
\BIBentrySTDinterwordspacing

\bibitem{R10}
N.~Jaques, S.~Taylor, A.~Sano, R.~Picard \emph{et~al.}, ``Predicting
  tomorrow’s mood, health, and stress level using personalized multitask
  learning and domain adaptation,'' in \emph{IJCAI 2017 Workshop on artificial
  intelligence in affective computing}.\hskip 1em plus 0.5em minus 0.4em\relax
  PMLR, 2017, pp. 17--33.

\bibitem{CNN}
\BIBentryALTinterwordspacing
T.~Song, G.~Lu, and J.~Yan, ``Emotion recognition based on physiological
  signals using convolution neural networks,'' in \emph{Proceedings of the 2020
  12th International Conference on Machine Learning and Computing}, ser. ICMLC
  2020.\hskip 1em plus 0.5em minus 0.4em\relax New York, NY, USA: Association
  for Computing Machinery, 2020, p. 161–165. [Online]. Available:
  \url{https://doi.org/10.1145/3383972.3384003}
\BIBentrySTDinterwordspacing

\bibitem{ROL}
W.~Liu, L.~Zhang, D.~Tao, and J.~Cheng, ``Reinforcement online learning for
  emotion prediction by using physiological signals,'' \emph{Pattern
  Recognition Letters}, vol. 107, 06 2017.

\bibitem{ML1}
K.~P. Murphy, \emph{Machine learning: a probabilistic perspective}.\hskip 1em
  plus 0.5em minus 0.4em\relax MIT press, 2012.

\bibitem{NN}
D.~Wang and Y.~Shang, ``Modeling physiological data with deep belief
  networks,'' \emph{International journal of information and education
  technology (IJIET)}, vol.~3, pp. 505--511, 01 2013.

\bibitem{DBN1}
G.~E. Hinton and R.~R. Salakhutdinov, ``Reducing the dimensionality of data
  with neural networks,'' \emph{Science}, vol. 313, no. 5786, pp. 504--507,
  2006.

\bibitem{SNAPSHOT}
\BIBentryALTinterwordspacing
A.~Sano, A.~Z. Yu, A.~W. McHill, A.~J.~K. Phillips, S.~Taylor, N.~Jaques, E.~B.
  Klerman, and R.~W. Picard, ``Prediction of happy-sad mood from daily
  behaviors and previous sleep history,'' in \emph{37th Annual International
  Conference of the {IEEE} Engineering in Medicine and Biology Society, {EMBC}
  2015, Milan, Italy, August 25-29, 2015}.\hskip 1em plus 0.5em minus
  0.4em\relax {IEEE}, 2015, pp. 6796--6799. [Online]. Available:
  \url{https://doi.org/10.1109/EMBC.2015.7319954}
\BIBentrySTDinterwordspacing

\end{thebibliography}

\end{document}